\documentclass{article}

\usepackage[square,sort,comma,numbers]{natbib}
\usepackage[final]{nips_2017}
\usepackage{graphicx}

\usepackage[utf8]{inputenc} % allow utf-8 input
\usepackage[T1]{fontenc}    % use 8-bit T1 fonts
\usepackage{hyperref}       % hyperlinks
\usepackage{url}            % simple URL typesetting
\usepackage{booktabs}       % professional-quality tables
\usepackage{amsfonts}       % blackboard math symbols
\usepackage{nicefrac}       % compact symbols for 1/2, etc.
\usepackage{microtype}      % microtypography
\usepackage{natbib}      % microtypography

\title{Prediction Scores as a Window into Classifier Behavior}

\author{
  Medha Katehara\\
  LNM Institute of Information Technology, India\\
 \texttt{medhakat95@gmail.com}\\
  \And
  Emma Beauxis-Aussalet \\
  Centrum Wiskunde and Informatica, Netherlands \\
  \texttt{Emmanuelle.Beauxis-Aussalet@cwi.nl} \\
\And
  Bilal Alsallakh \\
  Bosch Research, Palo Alto, USA \\
  \texttt{bilal.alsallakh@us.bosch.com} \\	
}

\begin{document}

\maketitle

\begin{abstract}

  Most multi-class classifiers make their prediction for a test sample by scoring the classes and selecting the one with the highest score.
	Analyzing these prediction scores is useful to understand the classifier behavior and to assess its reliability.
	We present an interactive visualization that facilitates per-class analysis of these scores.
	Our system, called Classilist, enables relating these scores to the classification correctness and to the underlying samples and their features. 
	We illustrate how such analysis reveals varying behavior of different classifiers.
	Classilist is available for use online, along with source code, video tutorials, and plugins for R, RapidMiner, and KNIME at \url{https://katehara.github.io/classilist-site/}.

\end{abstract}

\section{Motivation}

The function of a multi-class classifier is to classify an input sample into one out of many classes.
Prediction scores, when normalized, represent the probability of this sample to belong to each of the classes.
Having the sample with the highest score being consistently equal to the prediction target is the ultimate goal of classifier designers.
Consequently, the other scores computed by the classifier are often ignored, especially when the classifier is allowed to make one guess only.
This information, however, can be very valuable to understand the behavior of the classifier and to potentially optimize its design.
For example, Hinton et al. \cite{hinton2015distilling} demonstrate how these scores can be used as `soft-targets" to distill the knowledge from a large model into  a more compact one.
Viegas and Wattenberg demonstrate how the prediction scores can also reveal quality issues in the training and testing datasets \cite{Viegas2015Visualization}.
Likewise, \emph{Blocks} enables filtering confusion matrices based on prediction scores to reveal mislabeled samples or samples that fit in multiple valid targets \cite{alsallakh2017convolutional}.

A number of visualization systems have been proposed to analyze prediction scores using histogram plots.
Two notable examples are the Confusion Wheel \cite{alsallakh2014visual} and Squares \cite{ren2017squares}.
These plots were shown useful in understanding the behavior of the classifier.

We present a web-based system, called Classlist, to visualize prediction scores in relation with the classification correctness.
Our system combines selected design aspects from the above-mentioned systems to create a visualization that can handle tens of classes and, in the same time, is easy to interpret and interact with.
The next section briefly illustrates the main interface of Classlist.
Section~\ref{sec:example} provides examples insights about various classifiers that can be gained through Classlist.
Finally,  Section~\ref{sec:discussion} compares our system with previous ones, and elaborates on possible avenues for future work.

\section{Visualizing Prediction Scores using Classilist}

\begin{figure}
	\includegraphics[width=\textwidth]{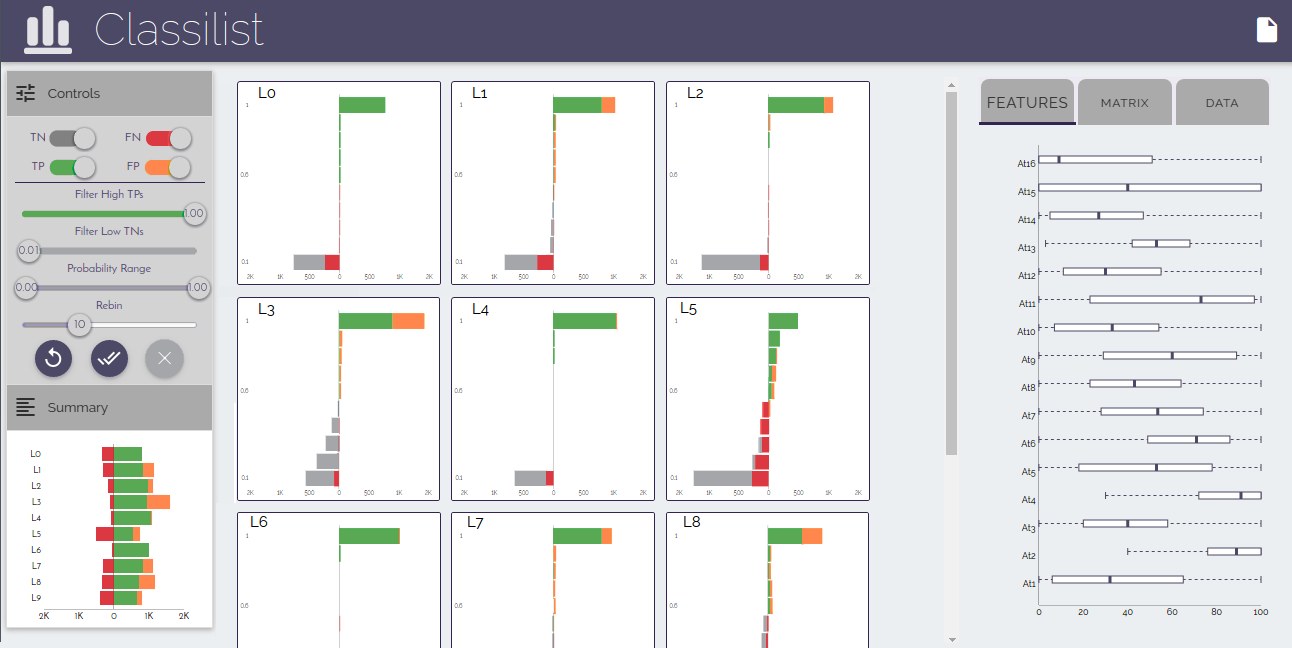}
  \centering
  \caption{A screenshot of Classilist showing classification results of the UCI Handwritten Digits dataset.
					The central view shows how the classification scores are distributed for each of the classes.
					Green, orange, and red indicate true positives, false positives, and false negatives respectively.
					The panel on the left enables filtering the samples by various criteria.
					The panel on the right enables exploring selected samples and their data features.}
	\label{fig:mainscreen}						
\end{figure}

Figure \ref{fig:mainscreen} depicts the main interface of Classilist showing classification results of the UCI handwriting digits dataset \cite{asuncion2007uci}.
This dataset contains 10,992 labeled samples representing handwritten digits.
A histogram is computed for each digit class to show the distribution of the corresponding scores, computed from all samples.
These histograms are shown in the central view.
The left panel enables the user to filter the samples in order to retain relevant ones only.
By default, the histograms contain samples that are predicted to be in the respective class, or that actually belong to the class. These samples encompass true positives (TPs), false positive (FPs), and false negatives (FNs), and are colored green, orange, and red, respectively.
As we show in section \ref{sec:example}, analyzing how each of these subgroups is distributed along the probability axis gives insight about the classifier's behavior and reliability.
A per-class selector of these groups is also available in the bottom of the left panel.

\paragraph{Additional Filtering}
The user can further include true negatives (TNs), especially those that are assigned non-zero scores to belong to the class.
Such samples represent borderline cases that exhibit high competition between two or more classes, and indicate possible risk of misclassification in case of slight changes in the input.
It is important to define a non-zero lower bound on the scores of the TNs to include, otherwise they might dominate the charts.
Users can further define an upper bound on TPs in order to focus on FPs and FNs, and to give them higher resolution in the charts.
Some classifiers produce low scores that do not span the whole probability range from 0 to 1.
In these cases, the user can adjust the probability axis to focus on the effective range, and then redo the histogram binning in order to increase the resolution of the visualization.
% of all samples in the test data along the probability range to belong the respective class.

\paragraph{Brushing and Details on Demand}

Classilist features three auxiliary views to explore classification data.
The first view visualizes the data features using box plots that show their value ranges.
The second view is a confusion matrix.
The third view is a sample viewer that lists individual samples and shows their features and pictorial representation if available.
Upon clicking on any bar in a probability histogram, the auxiliary views are updated to focus on the samples aggregated in this bar.
Furthermore, all occurrences of these samples in the other probability histograms and in the confusion matrix are highlighted.
For example, when selecting the FPs of a class, the corresponding FNs in other classes are highlighted.
Similarly, clicking on a cell in the confusion matrix highlights all occurrences of the respective samples in the histograms.
Such interactions are very useful to explore possible reasons behind misclassified samples.

% \cite{beauxis2014visualization}

\section{Example Insights}

\label{sec:example}

Inspecting and comparing the probability histograms across different classes or different classifiers reveals interesting variations among them.
These variations help to understand the behavior of different classifiers and in finding issues with the training data.
In the following we list three examples of insights we gained by analyzing different visual patterns.

\begin{figure}
	\includegraphics[width=\textwidth]{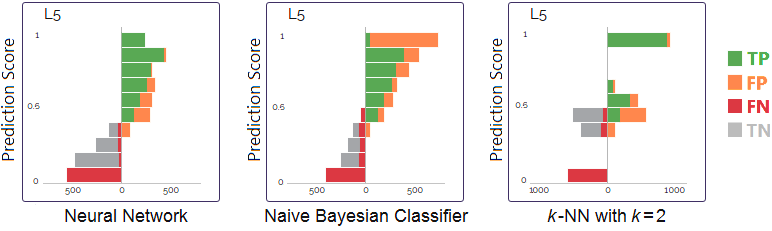}
  \centering
  \caption{The scores computed for all samples to belong to the class of "5" using three different classifiers (TNs with zero scores are filtered out).
	It is noticeable that the \emph{k}-NN classifier results in k+1 peaks in the histogram, while the other two result in more continuous distribution.
	Interestingly, the Naive Bayesian Classifier makes a lot of false predictions despite high prediction scores.
	}
	\label{fig:compare}
\end{figure}

\paragraph{Histogram Shape} 
We found that the overall shape of the histograms changes when using different classifiers.
Figure \ref{fig:compare} shows the histogram of scores computed for digit "5" when classifying the UCI handwritten digit using three different classifiers.
It is remarkable that the \emph{k}-Nearest-Neighbor classifier with $k = 2$ results in three peaks in the histogram.
The top-most peak represents samples whose two nearest neighbors are both samples of digit "5", and were hence predicted with very high probability to be of digit "5".
The middle peak represents samples whose nearest neighbor is a sample of "5" and second nearest neighbor is a sample of a different class, and were hence predicted with probablity around $50\%$ to be "5" .
The bottom-most peak represents samples of "5" that were misclassified as none of their two nearest neighbors were samples of "5".

\paragraph{Distribution of False Positives}
When a classifier makes a prediction with high probability, we expect it to likely be a correct prediction.
By looking at the probability histograms in Figure~\ref{fig:compare}, we can find out that not all classifiers have this property.
This is evident by looking at the distribution of false positives computed by the Naive Bayesian classifier, which surprisingly dominate the top-most bar of the histogram.
This is likely due to the importance of considering the joint distribution of point positions when classifying handwritten digits, which is discarded by this classifier.
In contrast, the \emph{k}-NN classifier makes very few errors when the prediction scores are high, as it accumulates more evidence to back these predictions.
Likewise, the error rate among the predictions of neural-network classifier increases gradually when the prediction score decreases. 

\paragraph{Biases in the Error Distribution}
In many case, the classifier exhibits biases towards or against certain classes.
While the confusion matrix gives a rough idea about these biases, the probability histograms provide further details.
For example we can see in Figure \ref{fig:mainscreen} that the classifier assigns generally low classification scores to samples of class "5", even when they are correctly classified.
In contrast, the classifier assigns very high scores to samples it recognizes of class "4".
This is in part because these samples have distinct shapes, compared with samples of other digits.
Interestingly, almost all of the samples that received scores in the range $[20\%, 40\%]$ for class "5" were actually of this class, but were still misclassified.
Adjusting these scores with a weighting factor would remedy these samples without incurring new false positives.
This indicates a clear bias against class "5", mostly towards class "3".
One reason behind this is the variation exhibited by the samples of digit "5" in the dataset, which results in smaller support for each variant during training.

These insights are hard to gain without inspecting the prediction scores.
This demonstrate the value of visualization in understanding classifiers' behavior.

\section{Discussion}

\label{sec:discussion}	

Classilist can handle a large number of samples, as these are binned in histograms.
The central view can display about 30 histograms at once on standard displays.
Users can still explore datasets with a larger number of classes by scrolling the view.

Classilist shows the same histogram information as in the Confusion Wheel \cite{alsallakh2014visual}.
It simplifies the visualization by using standard bar charts to show this information, instead of rotated ones.
It can handle more classes than Squares \cite{ren2017squares} by (1) placing the histograms in multiple rows within a scrollable panel, and (2) by avoiding a per-class breakdown of misclassified samples.
Such a breakdown requires a distinct categorical color for each class and results in a cluttered visualization.
Classee \cite{beauxis2015Bridging} offers an in-between solution, with a breakdown showing only the two classes yielding most FP and FN, and distinguished using shades of the same color.

Classilist is open-source and web-based. This aims to make the system accessible on several platforms without installation effort.
It can load the scores computed by any classifier along with the data features, as long as they are provided in the expected format.
We provided plug-in nodes for RapidMiner and KNIME that can be added to machine-learning pipelines in order to collect and export the data in the right format.
We also provide an R function for the same purpose.
Video tutorials on how to use these plug-ins as well as on online demo of our system are available at \url{https://katehara.github.io/classilist-site/}.

% This work was conducted as part of the Google Summer of Code 2016 program.

\bibliographystyle{abbrv}
\bibliography{references}

\end{document}